# Energy-Efficient Stochastic Computing (SC) Neural Networks for Internet of Things Devices with Layer-Wise Adjustable Sequence Length (ASL)

Ziheng Wang, Pedro Reviriego, Farzad Niknia, Zhen Gao, Javier Conde, Shanshan Liu and Fabrizio Lombardi

*Abstract*— Stochastic computing (SC) has emerged as an efficient low-power alternative for deploying neural networks (NNs) in resource-limited scenarios, such as the Internet of Things (IoT). By encoding values as serial bitstreams, SC significantly reduces energy dissipation compared to conventional floating-point (FP) designs; however, further improvement of layer-wise mixed-precision implementation for SC remains unexplored. This paper introduces Adjustable Sequence Length (ASL), a novel scheme that applies mixed-precision concepts specifically to SC NNs. By introducing an operator-norm–based theoretical model, this paper shows that truncation noise can cumulatively propagate through the layers by the estimated amplification factors. An extended sensitivity analysis is presented, using Random Forest (RF) regression to evaluate multi-layer truncation effects and validate the alignment of theoretical predictions with practical network behaviors. To accommodate different application scenarios, this paper proposes two truncation strategies (coarse-grained and fine-grained), which apply diverse sequence length configurations at each layer. Evaluations on a pipelined SC MLP synthesized at 32 nm demonstrate that ASL can reduce energy and latency overheads by up to over 60% with negligible accuracy loss. It confirms the feasibility of the ASL scheme for IoT applications and highlights the distinct advantages of mixed-precision truncation in SC designs.

## I. INTRODUCTION

Neural networks (NNs) are being increasingly recognized for their capability to effectively model and solve intricate problems, with widespread application in domains including robot control [1], recommender systems [2], natural language processing [3] and the Internet of Things (IoT). While efficient parallelized computations of NNs have been extensively investigated, they often come with stringent hardware complexity at nanoscales. Stochastic computing (SC) has emerged as an efficient hardware solution at low power for NN implementation [4]. SC NNs are particularly appealing for energy-constrained platforms due to the low overhead and excellent error tolerance. They also offer distinct benefits for specific applications, such as online learning and adaptive inference fine-tuning [4].

The computational scale of NNs has kept increasing to billions of parameters as evidenced in recent machine learning (ML) applications [5]. The demands for these networks place a significant requirement on energy dissipation; for tasks with smaller scales, such as IoT devices, energy efficiency is also crucial due to hardware and power constraints [6]. ML models have been widely applied to emerging 6G IoTs and other high-frequency communications, raising extensive challenges in terms of computation overhead and energy consumption [7]-[9]. Therefore, IoT devices usually employ compact NNs using techniques such as quantization or adaptive resolution [10], [11]. TinyML has focused on running ML models on devices with limited resources [12], always seeking to balance performance and efficiency but considerably reducing accuracy. Other approaches have proposed a hybrid method, executing predictions on the device except when performance is not adequate, in which case it is run in the cloud with more complex versions of the model [13]. Additionally, these applications usually require a fast response to user interaction and real-time communication with cloud or parallel devices, so low-latency hardware implementations are also needed [14].

Due to their efficient arithmetic features, SC NNs are affected by the stringent trade-off between accuracy and computational energy/latency [15]. A sufficiently long sequence is critical for accurate propagation and gradient updates in the NN; however, increased sequence lengths result in higher energy usage and latency, with serial processing of SC bitstreams presenting an approximately proportional dependency.

This research was partially supported by the Spanish Agencia Estatal de Investigación under Grants FUN4DATE (PID2022-136684OB-C22) and SMARTY (PCI2024-153434), by TUCAN6-CM (TEC-2024/COM-460), funded by CM (ORDEN 5696/2024), by the European Commission through the Chips Act Joint Undertaking project SMARTY (Grant 101140087), and by NSF under Grant CCF-1953961 and Grant 1812467. *Corresponding author: Shanshan Liu (email: ssliu@uestc.edu.cn); co-corresponding author: Zhen Gao (email: zgao@tju.edu.cn).*

Ziheng Wang, Farzad Niknia and Fabrizio Lombardi are with Department of Electrical and Computer Engineering, Northeastern University, MA 02115, USA.

Pedro Reviriego and Javier Conde are with the ETSI de Telecomunicación, Universidad Politécnica de Madrid, 28040 Madrid, Spain.

Shanshan Liu is with the School of Information and Communication Engineering, University of Electronic Science and Technology of China, Chengdu, 611731, China.

Zhen Gao is with the School of Electrical and Information Engineering, Tianjin University, Tianjin, 300072, China.

A reduction of the overheads in SC NNs could involve an intuitive strategy, i.e., using shorter sequence lengths, but it is crucial to achieve this without significantly compromising accuracy. When considering a scheme of mixed-precision inference for traditional FP NNs, various layers perform unique functions; hence, their impact on the outputs could be different [16]. Also, research indicates that layers within NNs differ significantly in their ability to tolerate errors [17], presenting the diverse effects of quantization errors on layers. However, mixed-precision inference in SC, or layer-wise adjustable stochastic sequences, has not been studied; in particular, by employing appropriate truncation strategies, different sequence lengths across layers in SC NNs are expected to improve energy and latency with a small accuracy loss.

This paper proposes an efficient scheme for SC NNs using an Adjustable Sequence Length (ASL) that is amenable to IoT applications. By employing layer-wise adjustable sequence lengths, the scheme contributes to reducing overheads while keeping a satisfactory accuracy. With the design directly truncating Sobol sequences in stochastic number generators (SNGs), ASL avoids additional costs and possible correlation problems. This paper evaluates the inference of Multilayer Perceptrons (MLPs) as per their well-developed SC designs [18], [20]. The main contributions of this paper are:

1) Propose ASL as the first dedicated, layer-wise truncation scheme in SC NNs, so enabling mixed-precision inference with reduced design overhead and high flexibility.
2) Develop an operator-norm–based theoretical model for accumulated truncation errors and validate it with random-forest (RF)-based sensitivity experiments. Evaluate for the first time truncation noise propagation in multi-layer SC NNs.
3) Justify that Sobol sequences preserve low correlation under truncation better than LFSR-based SNGs, bridging a key gap in SC hardware design by ensuring consistent accuracy even with truncated sequence lengths.
4) Present two distinct truncation methods for ASL, a coarse-grained approach for direct deployment and a fine-grained approach leveraging grid search, for better configuring layer-wise truncation under different accuracy-energy/latency trade-off requirements.
5) Confirm that ASL achieves substantial energy and latency reductions (up to more than 60 % savings) with negligible accuracy loss by synthesizing a pipelined SC MLP at 32 nm. Establish practical feasibility for IoT and other resource-constrained applications.

The rest of the paper is organized as follows: Section II reviews fundamental SC concepts and highlights the design features relevant to the proposed ASL scheme. Section III presents the theoretical analysis, including the noise model, operator-norm–based amplification factors, and empirical sensitivity analysis. Section IV proposes the ASL scheme and details the ASL truncation strategies by discussing both the coarse-grained and fine-grained approaches. Section V evaluates the ASL scheme by comparing theoretical estimates against actual hardware synthesis results, and it demonstrates the method's effectiveness in reducing energy/latency overheads. Section VI discusses the contributions of this paper. Finally, this paper ends with the conclusion and future works in Section VII.

## II. Preliminaries

### A. Stochastic Computing

In SC, a binary bitstream is utilized to represent a number in conventional digital arithmetic. Consider a $L$-bit stochastic sequence $S = (S_1, S_2, \cdots, S_L)$; each bit follows an independent and identical Bernoulli distribution with the parameter $p$ that can be interpreted as the probability of observing a "1" in the sequence [15]. The total number of "1"s can be represented as the sum of $S_1 + S_2 + \cdots + S_L$ in a binomial distribution. The encoded value $s$ is represented by a given stochastic sequence and has two formats:

*1) Unipolar representation*: A $L$-bit stochastic sequence with $L'$ "1"s encodes a value of $L'/L = p$. It represents a value in the range of $[0, 1]$ with data precision of $L^{-1}$.

*2) Bipolar representation*: A $L$-bit stochastic sequence with $L'$ "1"s encodes a value of $s = (2L' - L)/L = 2p - 1$. It represents a value within $[-1, 1]$ with a data precision of $2L^{-1}$.

Therefore, longer sequence lengths permit a higher computation precision (as shown in Fig. 1). Also, the Extended Stochastic Logic (ESL) can be applied to substantially increase the value range of SC representations [18]. A number can be represented by the division of two stochastic sequences, so it will not be limited to [0, 1] in the unipolar representation or [-1, 1] in the bipolar representation. By using the sequences for computation, SC can significantly reduce the complexity of arithmetic circuits in hardware implementations; for example, as shown in Fig. 1, the product of two numbers is simply computed by an AND gate in the unipolar representation. Therefore, SC is a promising scheme to implement computation-intensive systems (e.g., NNs) in which conventional computations using floating-point (FP) formats dissipate large power/energy.

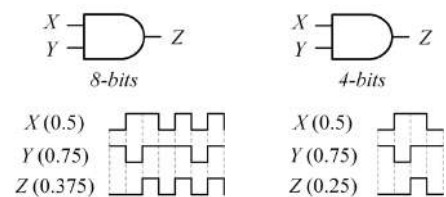

Fig. 1. Precision of unipolar multiplication with different stochastic sequence length (assuming low correlation between two sequences). A sufficient sequence length is critical to ensure an accurate computation.

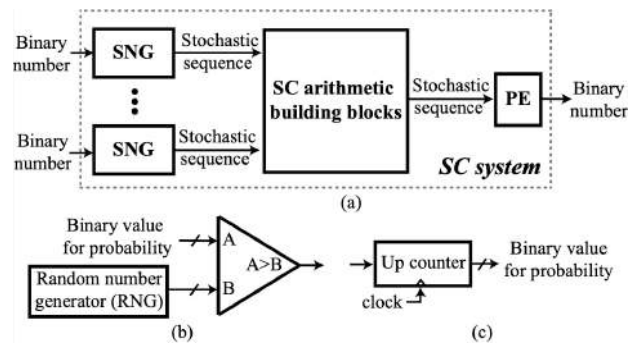

Fig. 2. Diagram of a generic SC system [34]: (a) the overall system; (b) a stochastic number generator (SNG); (c) a probability estimator (PE).

Fig. 2 shows the diagram of a generic SC system [34]. When implementing an SC NN, the main difference from the FP version is that conventional arithmetic modules are replaced by SC building blocks. Since this paper focuses on the inference process of NNs, the SC arithmetic building blocks shown in Fig. 2 (a) mostly include the units for implementing the multiplication-accumulation (MAC) process and the activation function. To implement the MAC, a network of XNOR gates is required to multiply the stochastic sequences generated for the inputs and their related weights; then, the adder tree is responsible for accumulating these SC products based on a multiplexer tree. A stochastic approximation to Tanh (STanh) is utilized after the ESL-based MAC process as the activation function. The above arithmetic units have been proven efficient for SC NN designs, and their implementation details are described in related works [18], [22], [23].

In addition to the arithmetic modules, the conversion circuits must be included in the SC design. As shown in Fig. 2, an SNG is essential for converting input data into stochastic sequences, and a probability estimator is used for converting back the sequences. The SNG is generally made of a random number generator (RNG) and a comparator. The RNGs can be implemented by a pseudo-random linear feedback shift register (LFSR) or quasi-random Sobol sequences [21]. Within the proposed ASL scheme, the selection of proper RNGs is crucial to ensure low correlation among truncated sequences. The Sobol sequence is preferred since it maintains a low correlation with truncation (as discussed in the subsequent sections). Further implementation details of the SC NNs and SNGs considered in this paper are presented in Section V.

*B. Impact of Sequence Length on SC NNs*

As per the arithmetic features introduced previously, the performance of SC designs is closely related to the sequence lengths. The error of an SC implementation originates from multiple sources: quantization error, random fluctuations, and correlation. The sequence length directly affects the quantization error, which is the primary error source, thereby significantly influencing the precision of an encoded value.

Consider an SC MLP with 6 layers with a size of 784-1024-1024-512-256-10 for the dataset Fashion-MNIST using the designs of [25] as an example. We have presented a simulation for the given network structure; the results are presented in Table I, and they show the effects of the sequence length on an SC MLP by two metrics: the averaged mean squared error (MSE) and the accuracy loss; those metrics are calculated using the conventional FP version with the same network configurations as the baseline. The MSE measures the absolute error caused by SC arithmetic compared with a traditional FP implementation in the critical MAC units in forward propagation. The values of layer outputs are within the range of [-1.78, 1.82], and the MSE results are averaged over all layers. The inference accuracy on 10,000 images from the validation sets is also presented to measure the overall performance of the network. The results show that the low MSE of the SC version with large sequence lengths leads to similar accuracy as the single-precision FP version; the degradation becomes more obvious with smaller sequence lengths, highlighting the step from 256 to 128 bits, where the accuracy decreases by a factor of 5.

TABLE I
PERFORMANCE OF SC MLPs WITH DIFFERENT SEQUENCE LENGTH FOR FASHION-MNIST

| Sequence Length (bits) | Average MSE | Accuracy Loss | #Cycles | Energy (mJ) |
|---|---|---|---|---|
| 1024 | 9.76e-5 | 0.02% | 5125 | 7.63 |
| 512 | 1.38e-3 | 0.04% | 2565 | 3.82 |
| 256 | 4.82e-3 | 0.11% | 1285 | 1.91 |
| 128 | 1.03e-2 | 0.52% | 645 | 0.95 |
| 64 | 8.01e-2 | 1.01% | 325 | 0.48 |

Table I also shows overhead metrics, the energy and the number of cycles, for the SC MLP with Fashion MNIST operating at 1 GHz; the design is implemented using Verilog HDL and synthesized using Cadence Genus at a 32 nm technology node; the timing constraint is set to 1 ns. As per these results, large sequence lengths cause high energy dissipation and more cycles (or equivalently more latency) because the stochastic sequences are computed in a bit-by-bit fashion. Moreover, these overheads scale almost linearly with the sequence length, as shown in Table I; such dependency could be exploited in the proposed ASL scheme to achieve significant latency/energy savings.

*C. Existing Mixed-precision Quantization and Opportunities for SC NNs*

Typically, the use of longer sequence lengths (in SC) or more complex data formats (in conventional FP) achieves a higher computation precision, but the hardware overheads also increase. However, for some applications that can inherently tolerate a small deviation in values (such as NNs), high precision may not be required for all computation modules; an adjustable computation precision could be employed to reduce hardware overheads while retaining satisfactory system performance. This has been investigated for NNs with FP computation by employing mixed-precision quantization inference [26], [27]. However, the mathematical parameters search and quantization-aware re-training require huge computational resources; for example, over 30x the time of the original forward propagation is needed [26]. Therefore, those applications could be limited in a practical context, especially for resource-constrained IoT platforms.

Similar approaches have not been investigated for SC NNs to further improve hardware efficiency; moreover, they have a large potential to achieve significant advantages by considering the features of SC. From the data representation perspective, the adjustment of sequence length can be easily performed by truncation. Note that all bits in an SC sequence are equivalent, and the encoded value $s$ is not changed after truncation in terms of mathematical expectation. As per the Law of Large Numbers, when the sequence is sufficiently long, the truncated part (if randomly chosen) still reflects the proportion of "1"s in the entire sequence [19]. This means that any part of a random sequence contains "1"s and "0"s in a similar proportion, reflecting the statistical characteristics of the entire sequence; further analysis and verification of the impact of sequence truncation are provided in Section IV.E.

From the perspective of hardware implementation, the conversion between different data precision in SC (i.e., sequence truncation) does not require any additional conversion unit due to the reasons discussed above. Moreover, the SC arithmetic circuits can be reused when applying different sequence lengths.

These advantages make the mixed-precision quantization principle for SC NNs very flexible and efficient. Therefore, an ASL scheme is proposed in this paper based on mixed-precision quantization; its benefits are expected to further address the issue of NN implementations for energy-constrained/high-performance platforms in which a traditional SC design may still incur considerable energy dissipation or latency.

## III. THEORETICAL ANALYSIS

### A. SC Truncation Noise Model and Assumptions

In conventional FP NNs, the quantization errors caused by low bit-width are commonly modeled as additive noise, and their accumulation through the layers is analyzed accordingly [26]. Likewise, when considering the SC NNs with a bitstream truncation scheme, we also encounter random fluctuations introduced by shortened bitstreams; these fluctuations propagate through the network layers and may degrade accuracy. Although both scenarios generate noise, their mechanisms differ slightly: SC depends on statistical fluctuations due to insufficient sample sizes, while FP relies on increased quantization errors due to larger intervals. Unlike the FP format, which involves distinguishing between high/low bits or exponent-mantissa segments, SC encodes values using bitstreams in which every bit is equally significant. Once the bitstream length $L$ is reduced, the only consequence is an increase in output variance; the representation grows roughly in inverse proportion $Var \propto 1/L$, rather than dealing with the exponent shifting or the mantissa clipping in FP formats.

To quantitatively evaluate the accumulation of these fluctuations in a multi-layer network, we use a model-based method on local linearity and independent noise, this has also been applied to FP quantization analysis [26]. The local linearity is based on the small perturbation assumption, which states that any truncation or quantization noise remains sufficiently small compared to the typical neuron activations, making it feasible to approximate the activation function $f(\cdot)$ by a first-order Taylor expansion. The second assumption enables us to treat the errors as independent additive noise. If the correlation between bitstreams can be managed (by Sobol sequence or other random sources), the truncation-induced errors in different layers can be viewed as independent with zero means. Under these assumptions, we can apply the superposition principle to analyze the effects of the noises in each layer on the final output.

Consider the $k$-th fully connected layer with weight matrix $W_k$ and $b_k$ and activation function $f_k(\cdot)$. Its output is given by

$$y_k = f_k(W_k y_{k-1} + b_k). \quad (1)$$

If we only introduce a small truncation noise $r_{W_k}$ into the $k$-th layer's weights, yielding $\widehat{W_k} = W_k + r_{W_k}$, the updated output can be written as $\widehat{y_k} = f_k\left((W_k + r_{W_k})y_{k-1} + b_k\right)$. Provided that $r_{W_k}$ is sufficiently small, we can approximate it by

$$\widehat{y_k} \approx y_k + J_k(y_{k-1})\delta_k, \quad \delta_k \propto r_{W_k}, \quad (2)$$

where $J_k(\cdot)$ is the Jacobian evaluated at $W_k y_{k-1}$. When multiple layers have zero means, mutually independent noise terms $\{r_{W_i}\}$, the resulting perturbations to the final output $Z$ can be treated in an approximately additive fashion. In variance terms, we have

$$Var(\sum_i r_{Z_i}) = \sum_i Var(r_{Z_i}). \quad (3)$$

This is very much analogous to FP quantization, in which quantization errors are likewise modeled as additive noise. The key difference is that in SC, the truncation noise depends inversely on the bitstream length $L$, while in FP arithmetic, the reduction of the bit-width leads to an exponentially larger quantization step size.

### B. Layer Noise Amplification Capacity

Based on the above assumptions of small perturbations and independent noise, we can not only analyze noise accumulation throughout the network, but also further examine the amplification capacity of each layer during noise propagation. This paper uses the concept of the operator norm from adversarial robustness in NNs [28], [29]. Specifically, the $l_2$ operator norm of each layer's weight matrix $W_k$ quantifies the worst-case amplification factor that the layer can apply to an input vector, i.e.,

$$\|W_k\|_2 = \max_{x \neq 0} \frac{\|W_k x\|_2}{\|x\|_2}, \quad (4)$$

which is equivalent to the largest singular value of $W_k$. Under the assumptions in Section III.A, the SC truncation noise can be regarded as a slight deviation on a linear mapping; therefore, if the input is contaminated by noise with norm $\|\delta\|$; the worst-case scenario is that the layer amplifies it to $\|W_k\|\|\delta\|$. The $l_2$ operator norm provides a concise upper bound $F = \|W_k\|$ when measuring the amplification of the noise at a given layer.

The amplification capacity of the activation functions should also be considered. For the STanh (as applied in this paper and mathematically equivalent to Tanh), the worst-case amplification factor does not incur any additional increase (the maximum gradient value can be taken as 1 with inputs centered around 0); therefore, the single-layer amplification factor with STanh can still be represented as $F = \|W_k\|$.

In a multi-layer network, the amplification of noise in each layer occurs sequentially, such that the overall effect is approximately the product of the operator norms of the layers. Therefore, if the truncation noise $r_{W_i}$ is injected at the $i$-th layer, the amplification upper bound from the $i$-th layer to the output layer $K$ is approximately $F_A = \prod_{k=i}^{K} \|W_k\|$. As per the SC scenario, the magnitude of the truncation noise $r_{W_i}$ scales as $\sim 1/L$ with the sequence length $L$. The theoretical upper bound on the noise amplitude at the final output can be expressed as

$$\|r_{Z_i}\| \lesssim \frac{1}{L} \prod_{k=i}^{K} \|W_k\|. \tag{5}$$

If one or more layers exhibit large $\|W_k\|$ values, even a slight truncation noise introduced in the early layers can be repeatedly amplified by the subsequent layers, resulting in significant errors in the output. Therefore, layers with larger accumulated amplification factors are more sensitive to SC truncation and they generally require longer bitstreams.

According to the analysis, we present theoretical results of each layer's amplification factors ($F$) as well as their cumulative effect ($F_A$) when cascaded towards outputs in Table II. The simulations in this paper apply to SC MLPs the datasets Fashion-MNIST [31], SVHN [32], and CIFAR10 [33]. In this part, a 6-layer fully connected network is analyzed as an example, and the analysis focuses on the first 5 layers (excluding the output layer). More implementation details are provided in the evaluation (Section V).

The accumulated amplification factors indicate that if noise is injected at the very first layer (Layer1), it can be cumulatively amplified hundreds of times at the final output (in the worst case), whereas the accumulated amplification factor of subsequent layers (such as Layer5) is rather small. This suggests that the early layers of the network are more sensitive to input noise, so when implementing the SC truncation strategy, special attention should be paid to ensuring the stability of these early layers.

Although the theoretical analysis provides accumulated amplification factors for noises occurring in each layer, these are merely worst-case estimates based on idealized assumptions. In practical applications, the sensitivity of each layer to SC truncation is often affected by many non-ideal conditions. Therefore, the next subsection focuses on an experimental analysis to evaluate the sensitivity of each layer and compare it with the theoretical results.

TABLE II
THEORETICAL RESULTS OF AMPLIFICATION FACTORS $F$ AND ACCUMULATED AMPLIFICATION FACTORS $F_A$ IN SC MLP

| Datasets | | Layer1 | Layer2 | Layer3 | Layer4 | Layer5 |
|---|---|---|---|---|---|---|
| Fashion MNIST | $F$ | 3.60 | 2.20 | 2.19 | 2.49 | 2.63 |
| | $F_A$ | 114.54 | 31.75 | 14.47 | 6.58 | 2.63 |
| SVHN | $F$ | 4.63 | 4.55 | 3.26 | 2.55 | 2.39 |
| | $F_A$ | 422.83 | 91.22 | 20.01 | 6.13 | 2.39 |
| CIFAR10 | $F$ | 4.05 | 3.64 | 2.76 | 2.47 | 2.86 |
| | $F_A$ | 290.10 | 71.59 | 19.61 | 7.09 | 2.86 |

*C. Sensitivity Analyses*

To quantitatively measure which layer is more sensitive to SC truncations, we refer to the approach in [26]: inject varying strengths of truncation noise solely into one layer (the $i$-th layer), then measure the accuracy drop $\Delta_{acc}$ and record the resulting perturbation $r_{Z_i}$ in the final output vector $y_k$. Under the small-noise assumption in (2), $\widehat{y_k} \approx y_k + r_{Z_i}$, then the magnitude $\|r_{Z_i}\|$ can be correlated with the observed $\Delta_{acc}$. When $\Delta_{acc}$ reaches a chosen threshold, we identify the sensitivity (or defined as "importance") of that particular layer. If a relatively small noise $\|r_{Z_i}\|$ can trigger a noticeable accuracy drop, it implies that this layer notably amplifies errors through subsequent propagation and thus, it warrants higher precision. Conversely, a layer capable of tolerating a larger $\|r_{Z_i}\|$ without major performance degradation allows for more aggressive truncation.

However, the method in [26] is limited because it only measures the effect when a single layer is truncated. Therefore, in this paper, we extend this approach using a more comprehensive method. The simulation uses the ASL scheme and grid search over all proper configurations (combinations of sequence lengths) by recording their accuracy. In this case, the effects of simultaneously truncating multiple layers can be clearly reflected in the output $\Delta_{acc}$. Different from directly looking at the importance by $\|r_{Z_i}\|$, our approach involves multiple factors; therefore, a regression model using an RF is applied to evaluate the effect of the sequence length of each layer on the inference accuracy (or the importance of each layer).

For comparison purposes, the network structure and configurations are identical to the ones applied in Section III.B. The configuration of sequence lengths can be denoted as $[L_1, L_2, \cdots, L_{k-1}]$, $L_i < L$; for example, with the full sequence length of $L = 2^{10}$, a proper range of truncated lengths can be set to vary as $L_i \in [2^6, 2^{10}]$.

In the RF model, the input features are the lengths of each layer, and the target value is the corresponding accuracy. With 100 trees, the importance of each feature (sequence length of different layers) is evaluated by its contribution to entropy. For example, if the truncation in the $i$-th layer leads to more reduction in the entropy for most tree nodes than other layers, then it is defined to have a so-called larger importance factor. The importance of all layers is normalized so that their sum is equal to 1; hence, we can get the importance of each feature (the sensitivity of the layers in terms of sequence lengths) as shown in Fig. 3. It shows the impact of truncation at different layers on the inference accuracy; a higher importance reflects that the truncation of the specified layer potentially leads to a larger accuracy degradation. Compared to the method in [26], which injects noise through truncation in only a single layer, the proposed RF method not only reveals the isolated impact of truncation on an individual layer, but also learns the cumulative contribution of each layer to the final accuracy under simultaneous multi-layer truncation scenarios.

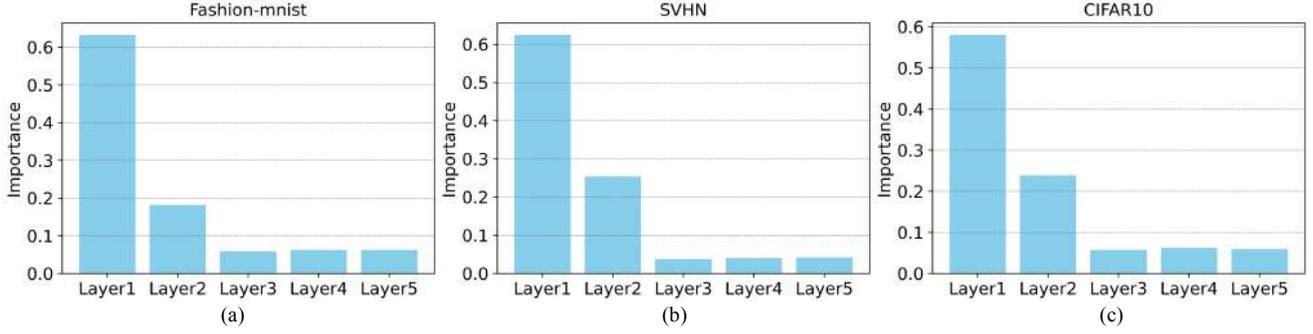

Fig. 3. The importance of the sequence length of each layer in a SC MLP on the inference accuracy. The results are evaluated by random forest on grid search over all the possible configurations with dataset: (a) Fashion-MNIST; (b) SVHN; (c) CIFAR10.

The results show that the sequence length of the early layers contributes the most to the inference accuracy. We have also simulated different networks and layer sizes of SC MLPs, and there are no obvious changes in the above conclusion. The first two layers contribute more than 80% to the inference accuracy; this indicates that the network is more sensitive to truncation in early layers. The conclusion is consistent with the theoretical analysis in Section III.B, showing that early layers are more sensitive to truncation due to accumulation effects. By the results in Table II, we can estimate the "theoretical importance" of the $i$-th layer simply by $F_{A,i}/\sum_{k=i}^{K} F_{A,k}$. If we compare the results in Fig. 3 with the "theoretical importance" calculated by the accumulated amplification factor listed in Table II, the importance of the first layer for Fashion-MNIST (66.45%) is similar to the theoretical estimation (67.38%) in Table II; while that importance for SVHN/CIFAR10 (65.33%/57.81%) is lower than the calculated values in Table II (77.92%/74.14%). The difference indicates that non-ideal conditions in practice do not perfectly follow the assumptions in theory. Also, in extreme cases, such as applying very short sequence lengths (< 64 bits), the small-perturbation assumption and linear approximations may not be valid (even though these setting are uncommon in practice due to severe accuracy loss). Therefore, the next subsection provides an experiment-based analysis; it is necessary to design proper truncation strategies based on both theoretical and empirical sensitivity analysis.

### D. Savings in Latency and Energy

Savings in latency and energy are also critical factors when selecting the truncation strategy. This part estimates the savings in these overheads by applying the proposed ASL scheme. As discussed in Section II.B, the latency and energy are approximately proportional to the sequence length in the SC implementations. In a pipelined architecture, MAC units and the sequence generation by SNGs operate concurrently, so allowing the analysis to concentrate only on the changes within the sequence generation process. As for latency, it is not related to layer size, because all computations within a layer are performed in parallel. By contrast, energy consumption correlates with layer size due to the computational load expanding with more neurons. The saving in energy $Saving_E$ (latency, $Saving_L$) can be approximated by the ratio of energy (latency) for the full model $Energy_F$ ($Latency_F$) and the ASL version $Energy_{ASL}$ ($L_{ASL}$):

$$Saving_L = 1 - \frac{Latency_{ASL}}{Latency_F} = 1 - \frac{\sum_{i=1}^{k-1} L_i}{(k-1)L}, \quad (6)$$

$$Saving_E = 1 - \frac{Energy_{ASL}}{Energy_F} = 1 - \frac{\sum_{i=1}^{k-1} L_i n_i n_{i+1}}{\sum_{i=1}^{k-1} L n_i n_{i+1}}, \quad (7)$$

where $L$ denotes the original sequence lengths of a $k$-layer full model (identical for all the layers) and $L_i$ is the sequence lengths of the corresponding truncated layers. $n_i$ is the feature size of the $i$-th layer in the network.

By the above equations, the expected savings in latency and energy are both related to the sum of the sequence lengths of all layers. The strict trade-off mentioned in Section II.B limits the use of small sequence lengths, so it is critical to arrange the configuration in each layer. Furthermore, the saving in energy benefits more from truncating the sequences in layers of large size, so those layers are more worthy of truncation. Note that (6) and (7) only consider the savings by layer sizes without extra factors such as hardware optimization and parallelization overheads. It only serves as a baseline calculation, while Section V compares the theoretical results (Table III) with an actual hardware evaluation (Table IV). This demonstrates how practical designs may introduce slight deviations from the ideal theoretical savings.

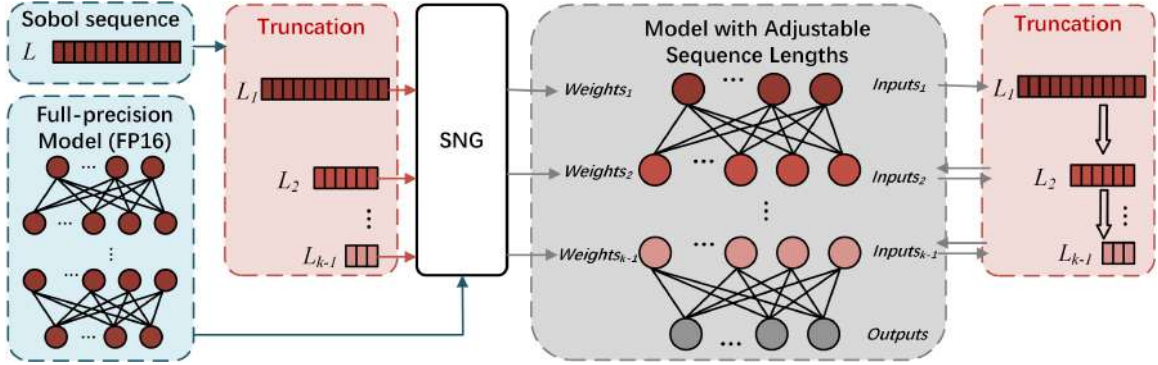

Fig. 4. The proposed ASL scheme with an FP16 full-precision base model. Weights are generated from truncated Sobol sequences, and the inputs of each layer are directly truncated to align with the format of the corresponding weights.

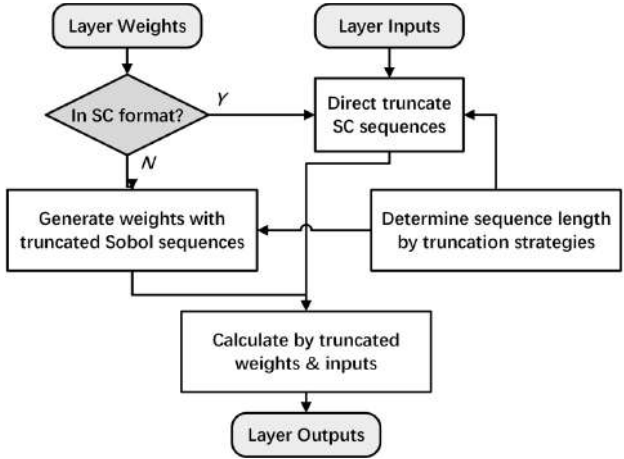

Fig. 5. Flowchart of the proposed ASL scheme in an SCNN layer.

## IV. PROPOSED ADJUSTABLE SEQUENCE LENGTH (ASL) SCHEME

This section first introduces in detail the proposed ASL scheme with its implementation steps. The truncation strategy for the proposed ASL scheme is then discussed, which is critical to assess the trade-off between accuracy and energy/latency.

### A. Overall Approach

The basic principle of the ASL scheme is to reduce the SC sequence lengths in the computational units; this can be realized by truncating the full model and inputs during inference.

Consider a full model of a $k$-layer SC MLP with sequence length $L$ as an example. The layer-wise computation includes the inputs of the $i$-th layer and the weight matrices between the $i$-th layer and the $(i + 1)$-th layer; the truncated sequence length of this computation is labeled as $L_i$. The output layer is not involved in computations, so the ASL scheme considers the first $k-1$ layers; the configuration of the sequence lengths of each layer can be represented as $[L_1, L_2, \cdots, L_{k-1}]$.

Since all bits in an SC sequence are equivalent, $L_i$ bits are randomly sampled from the original sequence of a full length $L$ to form the truncated sequence. The ASL scheme applies truncation to both the weight matrices and the inputs.

During inference, the inputs of each layer are directly truncated without regeneration. For the weight matrices, the truncation processes are slightly different when considering the format of the full model:
1) If the full model is stored in SC format, direct truncation can be applied to the weights.
2) If the full model is not stored in SC format, for example in single-precision FP format, the weights need to be converted into SC format. This can be realized by the SNGs using truncated Sobol sequences.

Directly truncating the generated SC sequences and truncating Sobol sequences before SNGs are mathematically equivalent; the use of truncated Sobol sequences requires lower memory and computational overheads. The forward propagation is then computed with the layer-wise truncated weights/inputs according to the configuration of the sequence lengths. The proposed ASL scheme is illustrated in Fig. 4, and the flowchart execution of employing it for a given layer is shown in Fig. 5. The truncation strategy, which is the critical part of the ASL scheme, is further discussed in the next subsection.

### B. Truncation Strategy: General Principles

The selection of the truncation strategies relies on distinguishing features of different layers in network propagation. Even though there is no related research on SC NNs, the strategy can partly refer to the studies in mixed-precision quantization for FP NNs; for example, [26] and [27] calculated the optimal bit-width of each layer by numerical methods. As described in the theoretical analysis in Section III, layer-wise quantization shares its core principle with the ASL scheme, as both approaches seek to reduce overheads by compromising on arithmetic precision. Their findings provide valuable insights; however, due to considerable differences in tasks and arithmetic (as mentioned in Section III), the truncation strategy for SC NNs requires both a dedicated analysis and simulation-based experiments.

The theoretical estimation in Section III.B and the empirical sensitivity analysis in Section III.C offer a clear view of the distinct characteristics of each layer in SC NN. It highlights the importance factor of the layers, showing the impact of their sequence lengths on the accuracy degradation. The conclusion is that the network is more sensitive to the truncation in early layers.

While the proposed operator norm analysis theoretically shows that the noise injected into early layers can be cumulatively amplified, additional structural and functional factors also reinforce this conclusion. Earlier layers typically carry out the essential feature extraction from the raw inputs: if these basic features are modified by truncation noise, subsequent layers will struggle to recover lost information. Moreover, many networks maintain a larger number of neurons or parameter sizes in earlier layers, allowing a stronger capacity for feature learning but also potentially creating a larger amplification factor. Later layers often have lower dimensions and operate on more abstract representations, hence being inherently less sensitive to truncation errors. Therefore, these factors in network design (core roles in feature extraction, differences in layer sizes, and hierarchical abstractions) converge with our theoretical estimate of accumulated noise amplification to explain the reasons by which early layers exhibit higher sensitivity to SC truncations.

According to the above discussions, the ASL scheme can apply a general consideration of truncation strategies: *Preserve the full sequence length in early layers (closest to the inputs) while truncating the latter layers (closer to the outputs).* This strategy focuses on reducing the overheads of SC NNs at a very much reduced accuracy loss compared to the full-precision model. Based on this principle, we introduce two distinct truncation strategies, a coarse-grained approach and a fine-grained approach, tailored to different application scenarios.

By the theoretical analysis in Section III, we can also intuitively infer how the truncation strategy should be modified with network architecture and data characteristics. Increasing the network depth generally amplifies truncation noise introduced in the early layers, due to the cumulative multiplication effect through successive operator norms. Likewise, higher data complexity (e.g., high-resolution images or multi-channel inputs) forces earlier layers to retain more detailed feature information. Consequently, in both scenarios, a more fine-grained scheme (that is discussed in the next subsections) is often needed to preserve enough precision.

In this paper, we only consider the implementation of ASL in small-scale NN architectures, because for large-scale models such as transformers SC implementations may remain uncommon due to stringent latency and hardware constraints. A full quantitative exploration of the guidance of complex network topologies and diverse data distributions on layer-wise truncation is left for future work.

### C. Coarse-grained Truncation Strategy

For the datasets and applications with no prior information, we can apply a coarse-grained truncation strategy; this applies generalized configurations that have satisfactory performance in most of the tasks. Based on the findings from the sensitivity analysis, the early layers of the network are critical for maintaining accuracy; therefore, to minimize degradation, truncation within the ASL scheme should primarily target the latter layers.

The use of the coarse-grained strategy depends on the general conclusion of sensitivity analysis in Section III.C. Assuming the original sequence lengths are equal $L$ in each layer of a baseline SC model, the ASL scheme keeps the first layer unchanged and truncates the remaining layers. The ratio of truncation can be approximately determined by the sensitivity analysis in Fig. 3, for example, the importance of the second layer falls in the range of $[1/4, 1/2]$ of the importance of the first layer. Since the length of the stochastic sequence is typically in the form of a power of 2, we select the factor of $1/2$ as a generalized truncation strategy (taking the upper bound as a safe choice to preserve accuracy). Similarly, we can further truncate the remaining layers by a similar approach, and the general strategy can be averaged considering the results from all the datasets. Therefore, the overall coarse-grained truncation strategy can be represented as $[L, L/2, L/4, \cdots, L/4]$. Once the sequence length $L$ is determined, the theoretical savings in energy and latency can be calculated by (6) and (7). The performance of this generalized ASL configuration is evaluated in Section V.

### D. Fine-grained Truncation Strategy

While the generalized truncation strategy is effective in many scenarios, it may not always provide an optimized balance between accuracy and energy/latency savings. To enhance performance, a fine-grained truncation approach can be applied. Testing a wide range of configurations on a large dataset can be costly, thus, employing a smaller, randomly selected subset is a practical solution. This subset can be treated as statistically representative of the entire dataset during inference, so allowing for the identification of consistent patterns. Cross-validation confirms that the fine-grained strategy is effective across both subsets and the complete dataset.

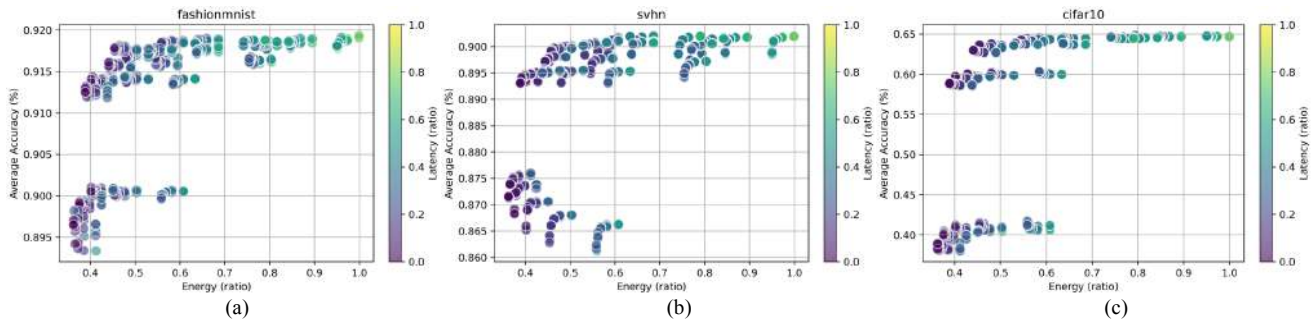

Fig.6. Illustration of the relationship between accuracy loss and savings (ratio) of energy and latency when applying the ASL scheme on dataset: (a) Fashion-MNIST; (b) SVHN; (c) CIFAR10. The points denote configurations with different truncation strategies.

Utilizing a grid search to explore all viable configurations of sequence lengths on the chosen subset allows for detailed tracking of the accuracy loss. The implementation of the sensitivity analysis as preliminary knowledge can enhance efficiency, such as by maintaining a constant sequence length for the critical first layer to reduce the computational demands in the grid search. For each configuration, the expected savings in energy and latency can be calculated by (6) and (7). The results are illustrated in Fig. 6, in which the point with 100% savings in both energy and latency indicates the baseline no accuracy loss. We can observe similar patterns showing the trade-off between accuracy loss and savings. Notably, certain configurations stand out by offering substantial savings while reducing the accuracy loss.

The score for the savings can be calculated as

$$score = \alpha Saving_E + (1-\alpha)\, Saving_L, \quad (8)$$

where $\alpha$ is the weight balancing the importance of the energy and latency savings; it can be determined by the requirement of specific implementations, for example, this paper applies $\alpha = 0.5$ for identical importance of the two terms. For a given threshold of accuracy loss ($\Delta_{acc} < 0.1\%$), it is possible to find the configuration corresponding to the point that achieves the highest score. Such configuration can be applied as the fine-grained strategy of the ASL scheme.

Therefore, the fine-grained strategy can be summarized in the following algorithmic steps:
1) Randomly select a subset from the given dataset.
2) Grid search by inferencing all possible configurations (with a proper range of truncated lengths such as $L_i \in [2^6, 2^{10}]$) over the selected subset and recording the corresponding accuracy loss. The conclusions from the sensitivity analysis can be applied to reduce candidate configurations to evaluate.
3) Calculate the expected savings in energy and latency by equations (6) and (7).
4) Determine the overall score of the configurations by equation (8).
5) For a given threshold of accuracy loss, find the configuration achieving the highest saving score.

According to previously presented simulation results, the configuration selected by such a process is expected to achieve a better performance than the coarse-grained strategy during the inference of full datasets. A detailed evaluation of the fine-grained truncation strategy is presented in Section V.

*E. Implementation of RNGs*

The ASL scheme can be realized by applying truncated Sobol sequences in the SNGs. This subsection details the implementation of the RNGs, hence justifying the use of Sobol sequences. To achieve accurate computation, SC arithmetic requires an adequate length for the stochastic sequences involved. Since the proposed ASL scheme relies on truncated stochastic sequences, low correlation must be preserved after truncations.

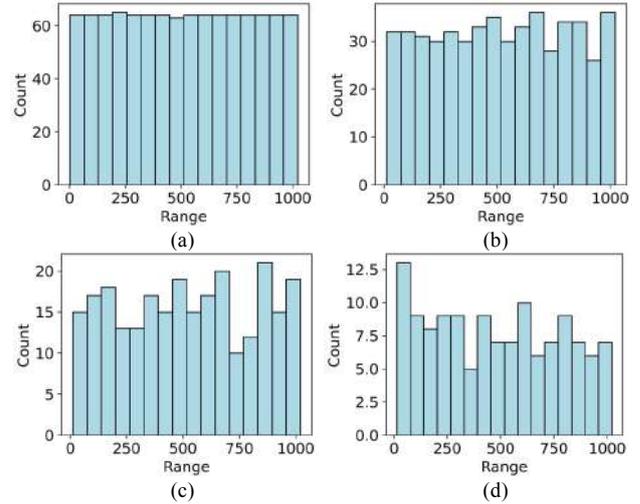

Fig. 7. Distribution of pseudo-random numbers generated by an LFSR truncated to a length of (a) 1024 (original); (b) 512; (c) 256; (d) 128.

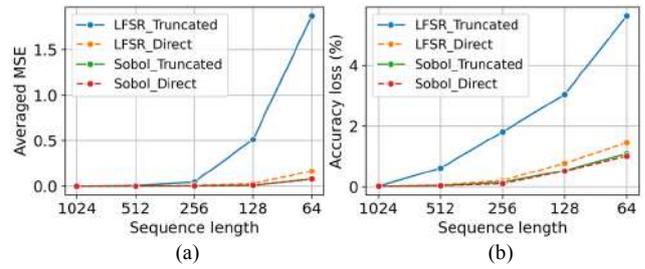

Fig. 8. Comparison of SC MLPs with truncated and directly generated sequences using LFSR or Sobol as RNGs: (a) Averaged MSE; (b) Accuracy Loss. Fashion-MNIST is used and the results are calculated using the single-precision FP implementation as baseline.

RNGs in SC NNs are usually implemented by LFSRs or Sobol sequences. An LFSR is a streamlined technique for pseudo-random number generation, utilizing a basic binary shift register and linear feedback rules based on a specific polynomial. Its simplicity allows for the fast generation of pseudo-random sequences with desirable statistical properties. As used for generating quasi-random sequences, Sobol sequences differ from LFSRs because they prioritize the uniform distribution across multi-dimensional spaces.

When implementing the ASL scheme with truncated stochastic sequences, the application of a LFSR is limited. As directly truncated stochastic sequences are equivalent to stochastic sequences produced from truncated random sequences, Fig. 7 provides its intuitive illustration: the original random numbers generated by the LFSR (with a length of 1024) exhibit satisfactory statistical properties (they can be guaranteed by well-selected polynomials and seeds). However, truncation disrupts the uniform distributions (Fig. 7 (b), (c), and (d)), with the deviation intensifying as the length of the truncated sequences decreases. By contrast, Sobol-generated SC sequences maintain a consistent advantage due to their strict inherent uniform distribution [21], irrespective of truncation; therefore, Sobol sequences are the preferred choice for RNGs in the ASL scheme.

Next, an evaluation by simulation can be used to intuitively compare the performance of those two cases of SC MLPs. The stochastic sequences generated by truncated LFSR/Sobol sequences are compared with those generated by directly produced LFSR/Sobol sequences, encoding identical values (the latter approach can employ proper seeds, so it can always ensure low correlation). Like in Section II.C, the MSE of an SC implementation in the critical MAC units and the accuracy loss of SC NN are evaluated by comparing them with the traditional single-precision FP implementation. The results for dataset "Fashion-MNIST" in an MLP with an identical configuration as in Section II.C are presented in Fig. 8 as an example (the original stochastic sequence length before truncation is 1024).

The results indicate that the LFSR version suffers a large difference between the truncated version and the directly generated version, confirming its inadequacy for the truncation scheme as previously concluded. However, the Sobol-based truncated SC sequences show MSE and inference accuracy levels comparable to those of the directly generated sequences (with the same length and encoded values). This confirms the feasibility of the proposed scheme: when truncated Sobol sequences are utilized in RNGs, the ASL emerges as an effective strategy, achieving negligible degradations in overall performance.

## V. PERFORMANCE EVALUATION

This section conducts simulations to assess the performance of the ASL scheme with datasets Fashion-MNIST [31], SVHN [32], and CIFAR10 [33] (one channel is used as input for each image in SVHN and CIFAR10). For each dataset, the scheme is evaluated by the validation set with 10,000 images. The SC MLPs are implemented based on the design of [25]; STanh is applied as the activation function, and MLPs with different sequence lengths are trained as baseline models. For the coarse-grained truncation strategy, this paper uses the configuration $[L, L/2, L/4, \cdots, L/4]$ as discussed in Section IV.III.

When applying the fine-grained truncation strategy and for comparison purposes, the threshold of the accuracy loss is set to be the same as the coarse-grained strategy. The configurations with the best performances are selected based on the pattern tested on a subset with 5% size of the original dataset. The detailed configurations of the sequence length in each layer are presented in Table III.

### A. Implementation of SC NNs

Before conducting the evaluation, the hardware implementation of SC NNs considered in this paper is described. As introduced in Section II.A, the implementation of SC NNs mainly includes the circuits for performing MAC and the activation function. Fig. 9 (a) shows the design for a layer propagation. Different from the direct inputs in traditional FP networks, the inputs of SC need SNGs to generate uncorrelated stochastic sequences. To implement the MAC, a network of XNOR gates is required to perform multiplication between the stochastic sequences generated for the inputs and their related weights; then, the adder tree accumulates these SC products based on a multiplexer tree; the adder tree contains a pipeline stage to meet the tight timing constraints. By moving forward in the adder tree, the number of adders in each level decreases; therefore, the pipeline stage is placed from the output side to decrease the complexity of the sequential logic.

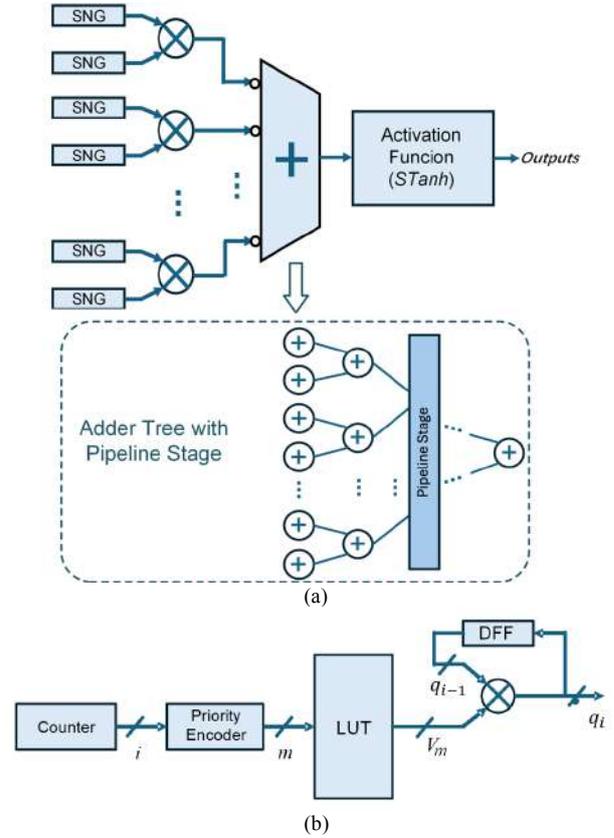

Fig. 9. The hardware design of an SC NN: (a) implementation of the layer propagation including SNGs, MAC with a pipelined adder tree and activation function; (b) the Sobol sequence generator (i.e., RNG) in an SNG.

For implementing the activation function, a stochastic approximation to Tanh (STanh) is utilized after the ESL-based MAC process. It is implemented using a LFSM as the design proposed in [23]. The use of the clock in the LFSM-based activation function introduces a further sequential logic in the design to perform as another level of pipelining.

For the SNG implementation, this paper applies the Sobol sequence to maintain a low correlation as discussed in Section IV.E. The RNG, i.e., the Sobol sequence generator, is designed according to the scheme presented in [21], [24]. As shown in Fig. 9 (b), its hardware design includes a counter for generating index $i$ and then a priority encoder for generating the signal $m$ according to the index and its leading zero position. Signal *m* represents the required vector's ($V_m$) address that is saved in the Look Up Table (LUT) beforehand. The final Sobol sequence is obtained after reading this vector and XORing it with the Sobol sequence generated in the previous cycle ($q_{i-1}$). Such Sobol sequence is then compared with the comparator in the SNG to generate the stochastic sequence.

The above units have been proven to be efficient for SC NN design, and their implementation details are described in related works [18], [22], [23]. Based on these units, a fully parallel network is implemented, i.e., the operation of each neuron in the network is independently performed, regardless of the layer. However, this does not imply that a neuron's value is generated in one cycle. Since in SC, the final value of a neuron is dependent on the sequence, the fully parallel implementation operates on a bit basis of the sequence per cycle; therefore, the total latency of the design is only related to the sequence length and independent of the network size due to the full parallelization. To address the critical path delay, if the number of neurons in the previous layer (input or hidden layer) is significantly large, the number of adder stages for the accumulation design of Fig. 9 (a) increases, this causes an increase in total critical path delay. To address this delay problem, a pipeline stage is inserted from the outer side of the accumulator; this scheme also reduces the number of registers, because the number of bits is reduced by moving toward the end of the adder stages. This scheme may increase the complexity of the design, but it improves its performance and can achieve a high operating frequency (i.e., 1 GHz in this paper). Further details of the parallel SC NN design with pipeline can be found in [25]. Such a design is taken as a baseline to conduct the hardware evaluation for the proposed ASL scheme in Section V.C.

In the hardware implementation, we employ both pipelining and parallel design so that each neuron's bit-serial operation can complete within the original 1 ns timing constraint. Specifically, even if the network scales up or the bitstream is truncated to a different length, the critical path does not grow beyond the synthesizable limit. Instead, the core logic of each neuron remains confined to one pipeline stage, thereby sustaining the same 1 ns clock period. This design ensures that the overall latency is linearly dependent on sequence length and is not affected by layer sizes; the design preserves sufficient timing margins under varying network configurations and truncation scenarios.

### B. Theoretical Savings in Latency and Energy

For the strategies proposed in the previous section, we obtained the coarse-grained and fine-grained configurations for SC MLPs with the proposed ASL scheme. For comparison purposes, we can first estimate the savings by the given strategies to confirm the theoretical analysis presented in Section III.D. Table III lists the estimated savings in energy and latency calculated by equations (6) and (7). The savings in latency are only dependent on the sequence lengths, irrespective of the layer size, making them consistent across various datasets with the same configuration. This consistency is evidenced by the coarse-grained strategy that delivers the same savings of 55%. The savings in energy are related to the layer sizes. Since for the proposed scheme it has been suggested to truncate the latter layers, then the networks with a smaller first layer tend to gain more benefits; hence, Dataset Fashion-MNIST achieves a larger energy saving (up to 40.6%) than dataset SVHN/CIFAR10 (up to 36.9%) with the same configuration. Moreover, the fine-grained strategy allows the use of small sequence lengths with specified datasets, while satisfying the accuracy threshold; hence, these configurations lead to larger savings in both energy and latency. As evaluated by a pipelined design, the savings in practice are further assessed in the next subsection.

### C. Performance and Hardware Evaluation

The mixed-precision quantization for the FP NNs has been widely studied and can achieve a significantly larger compression rate, such as sub-byte representations with an increased accuracy degradation [26], [27]. Given the differences between FP and SC in network implementations, a direct comparison between truncation strategies of those arithmetic is not feasible. While mixed-precision FP papers do not typically discuss hardware evaluations, they require additional overheads for format conversions, including bias adjustments in exponent bits and roundings in mantissa bits or a specially designed fused MAC [30]. These overheads become more pronounced with frequent layer-wise format conversions and larger network sizes. However, as outlined in Section II.C, mixed-precision SC strategies can be straightforwardly implemented by truncating the same stochastic sequences, thereby dramatically reducing the conversion overhead.

This section focuses on highlighting the benefits of mixed-precision techniques in SC and assessing the network performance with or without the proposed ASL scheme. A case study is presented in which the performance of ASL is evaluated by hardware design. For illustration purposes, we set the sequence lengths of the full models to $L = 1024$. A 1 GHz stochastic MLP [25] is designed using Verilog HDL and synthesized using Cadence Genus at a 32 nm technology node; the timing constraint is set to 1 ns. The results for the SC MLPs with the two ASL truncation strategies (with the detailed configurations listed in Table III) are presented in Table IV.

TABLE III
THEORETICAL SAVINGS IN ENERGY AND LATENCY BY THE ASL SCHEME (COMPARED TO FULL MODEL WITH $L$=1024, $K$=6).

| Dataset | Layer sizes | Truncation strategy | Configuration | %Savings in energy | %Savings in latency |
|---|---|---|---|---|---|
| Fashion-MNIST | 784-1024-1024-512-256-10 | Coarse-grained | $[2^{10}, 2^9, 2^8, 2^8, 2^8]$ | 40.6% | 55.0% |
| | | Fine-grained | $[2^{10}, 2^9, 2^7, 2^6, 2^6]$ | 44.2% | 65.0% |
| SVHN | 1024-1024-1024-512-256-10 | Coarse-grained | $[2^{10}, 2^9, 2^8, 2^8, 2^8]$ | 36.9% | 55.0% |
| | | Fine-grained | $[2^{10}, 2^9, 2^8, 2^6, 2^6]$ | 37.8% | 62.5% |
| CIFAR10 | 1024-1024-1024-512-256-10 | Coarse-grained | $[2^{10}, 2^9, 2^8, 2^8, 2^8]$ | 36.9% | 55.0% |
| | | Fine-grained | $[2^{10}, 2^9, 2^8, 2^7, 2^6]$ | 37.6% | 61.3% |

TABLE IV
PERFORMANCE OF THE ASL SCHEME IN HARDWARE EVALUATION WITH DIFFERENT CONFIGURATION (COMPARED TO FULL MODEL WITH $L$=1024).

| Dataset | Layer sizes | Baseline Accuracy | Truncation strategy | Energy (mJ) | #Cycles | Accuracy Loss | %Savings in energy | %Savings in latency |
|---|---|---|---|---|---|---|---|---|
| Fashion-MNIST | 784-1024-1024-512-256-10 | 91.98% | Coarse-grained | 4.08 | 2309 | 0.083% | 46.59% | 54.95% |
| | | | Fine-grained | 3.83 | 1797 | 0.083% | 49.83% | 64.94% |
| SVHN | 1024-1024-1024-512-256-10 | 90.42% | Coarse-grained | 4.75 | 2309 | 0.040% | 42.82% | 54.95% |
| | | | Fine-grained | 4.68 | 1925 | 0.040% | 43.64% | 62.44% |
| CIFAR10 | 1024-1024-1024-512-256-10 | 64.86% | Coarse-grained | 4.75 | 2309 | 0.098% | 42.82% | 54.95% |
| | | | Fine-grained | 4.70 | 1989 | 0.095% | 43.38% | 61.19% |

In our simulation, the delays are identical (1ns) for neurons with different SC lengths (as per timing constraint in the synthesis process), while the total number of clock cycles is proportional to the SC length; therefore, for the pipelined design with the same NN topology and truncation strategy, the percentage of savings in latency is identical, while the percentage of savings in energy is dependent on layer sizes. As per Table IV, the ASL scheme with a coarse-grained truncation strategy achieves energy savings of 46.59% (42.82%/42.82%) for the dataset Fashion-MNIST (SVHN/CIFAR10), and identical savings of 54.95% in latency across the datasets. The fine-grained strategy achieves larger savings, especially in latency, with a similar level of accuracy loss; it achieves energy savings of 49.83% (43.64%/43.38%) on Fashion-MNIST (SVHN/CIFAR10), and latency savings of 64.94% (62.44%/61.19%) on Fashion-MNIST (SVHN/CIFAR10). Overall, the synthesis results of latency are very close to the theoretical estimation in Table III (the differences are < 0.1%), while the savings in energy are larger than the estimated results. The main reason is that the power of layer propagation is not strictly linear to the layer sizes, leading to a deviation compared with the theory in Section III.D.

The accuracy of SC NNs when employing different strategies in the ASL scheme is also evaluated and compared. The simulation is performed in Pytorch using the functionally equivalent network implementation and results are also provided in Table IV. For the full model with a sequence length of $L = 1024$, the accuracy loss threshold is set to $\Delta_{acc} < 0.1\%$ when selecting the fine-grained strategies; this is negligible, especially compared with the inherent quantization (rounding) errors of SC formats.

Overall, as per the simulation results, the coarse-grained strategy should satisfy the requirement of most tasks; if further improvements are required, the fine-grained strategy can be applied with some additional costs in improving the configurations. In summary, the case study shows significant savings in overheads with negligible accuracy loss by applying the ASL scheme to SC MLPs, so confirming the efficiency of the proposed method in source-limited applications.

## VI. CONTRIBUTIONS

This paper introduces a novel layer-wise truncation ASL scheme. To the best of the authors' knowledge, there is no related research on mixed-precision SC implementations. The presented work also diverges from traditional FP mixed-precision methods by focusing on stochastic bitstreams, in which every bit is equally significant. Unlike FP quantization, which relies on discrete bit-width adjustments often demanding expensive exponent or mantissa manipulations, ASL simply truncates bitstreams in SC. It significantly reduces the energy/latency overhead and maintains adequate precision. This difference is particularly relevant to constrained scenarios such as IoT and further addresses the advantages of SC with area and power limitations.

Beyond these architectural differences, our work presents a unified theoretical and empirical framework. An operator-norm–based model captures worst-case noise accumulation across layers, highlighting the amplification of the truncation noise from early to deeper layers. The theory has then been validated with an improved RF-based sensitivity analysis that covers multi-layer truncation scenarios. By incorporating Sobol sequences instead of LFSRs, the proposed scheme ensures that truncated bitstreams retain a low correlation, which is an important challenge that prior SC works had not thoroughly addressed.

From a practical application perspective, our results show that the proposed coarse-grained and fine-grained truncation schemes achieve over 60% energy and latency savings with negligible accuracy loss, validated on hardware synthesized at the 32 nm technology node. As IoT devices require small overhead and real-time responsiveness, the ability to adjust sequence lengths on a layer-wise basis offers a powerful approach for balancing efficiency and accuracy. This further facilitates the use of SC NNs in edge and embedded systems in which conventional FP hardware proves costly. By theoretical modeling and hardware evaluation, this paper offers an effective approach toward high-performance, low-power SC implementations suitable for next-generation IoT and other resource-constrained ML applications.

## VII. Conclusion and Future Work

This paper has introduced the ASL scheme for SC NNs, a novel layer-wise truncation approach that significantly reduces the overhead while preserving high accuracy in resource-constrained platforms. Through theoretical analysis and an extended multi-layer sensitivity evaluation, we have shown that noise in early layers can be cumulatively amplified, thereby motivating the need for maintaining longer bitstreams at those layers.

This paper has further proposed two complementary strategies, a coarse-grained approach for general tasks and a fine-grained grid-search-based approach for scenarios requiring tighter trade-offs. By exploiting Sobol sequences, this paper has also addressed correlation issues in truncated bitstreams, presenting the scheme of layer-wise truncation for practical SC implementations. The hardware evaluation of a pipelined SC MLP has confirmed that ASL achieves significant energy and latency savings (exceeding 60% in certain cases) with negligible accuracy degradation.

Future work is directed to extend ASL to more complex architectures such as convolutional or recurrent NNs and explore adaptive or dynamic truncation strategies; incorporating techniques from traditional FP networks such as quantization-aware training can potentially further refine the SC truncation scheme. Additionally, investigating hardware-aware optimizations would enable a tighter integration of ASL into energy-limited devices. This can further address the advantages of SC NNs in IoTs and edge computing applications.